%% file: main.tex
\documentclass[letterpaper, 10 pt, conference]{ieeeconf}
\IEEEoverridecommandlockouts
\usepackage[english]{babel}
\usepackage[T1]{fontenc}
\usepackage{amsmath}
\usepackage{amsfonts}
\usepackage{bm}
\usepackage{mathtools}
\usepackage{graphicx}
\usepackage{xcolor}
\usepackage{booktabs}
\usepackage{array}
\usepackage{multirow}
\usepackage{subcaption}
\usepackage{tikz}
\usepackage{cite}
\usetikzlibrary{positioning,fit,arrows.meta,calc}
\input{shorthands/shorthands.tex}

\usepackage[hyphens]{url}
\usepackage[hidelinks]{hyperref}

\title{\LARGE \bf
Generalizing Manipulation Skills with a Local Coding Agent
}

\author{Raman Talwar$^{1*}$, Elias Nijs$^{1*}$, Andreas Verleysen$^{1}$ and Francis wyffels$^{1}$%
\thanks{$^{*}$Equal contribution.
$^{1}$IDLab-AIRO, Ghent University -- imec, Belgium.}%
}

\begin{document}
\maketitle
\thispagestyle{empty}
\pagestyle{empty}

\begin{abstract}

Today, progress in open-weight language models enables systems capable of
writing, executing and debugging code while still running on a single
workstation. Most language-driven robots give the model a fixed action
interface or a trained policy. Generalizing to a new task therefore
means more engineering effort or more data collection, both
time-consuming. We investigate whether a local open-weight
vision-language model can control a robot and one-shot generalize to
new variations of a task without new human programming or training. We
let a local open-weight VLM, Qwen3.8-27B, drive a UR3e robotic arm from
a coding-agent harness. It writes and runs its own code above a service
that implements kinematics, safety limits and classic computer vision
techniques. We investigate if this system is capable of generalizing to
unseen tasks. Specifically, we test it on nine tasks built from
children's toys designed to probe generalization capability across
various object characteristics: color, size, shape, and task variation
of those. With five trials for each task, we observe generalization in
30 out of 45 trials with durations ranging from 3.4 to 67.5 minutes
depending on task complexity. We further test if there is a speedup
when an agent is asked to redo the task after successful completion.
This resulted in a 50\% reduction in duration, indicating that there is
self-improvement over time. Finally, we expose the limitations of a
local coding agent. We believe that solving those limitations combined
with further investigation of self-improvement over time points at a
direct path toward real-world deployment of a local coding agent.

\end{abstract}

\input{sections/introduction.tex}
\input{sections/related.tex}
\input{sections/method.tex}
\input{sections/experiments.tex}
\input{sections/discussion.tex}

\input{sections/conclusion.tex}

\section*{ACKNOWLEDGMENT}
Parts of the manuscript text were drafted and revised with Claude Code (Fable 5.1, Opus 5.1). It was also used in drafting the skills for the framework and editing code.
All AI usage happened under the authors' direction. The authors checked every stated
number and claim.

This research was partially funded by the Flanders AI Research Program.

\bibliographystyle{IEEEtran}
\bibliography{frontback/bibliography}

\end{document}

%% file: shorthands/shorthands.tex
\definecolor{linkblue}{HTML}{0099cc}
\definecolor{linkpurple}{HTML}{9400D1}

\definecolor{mainred}{HTML}{ae4132}
\definecolor{mainblue}{HTML}{10739e}

\definecolor{framegray}{HTML}{8A8A8A}

\def\secref#1{section~\ref{#1}}

\def\eqref#1{equation~\ref{#1}}

\def\1{\bm{1}}


%% file: sections/introduction.tex
\section{INTRODUCTION}
\label{sec:intro}

Instructions in natural language are most of the time ambiguous for robots.
\emph{Stack the smaller cup on the bigger one} does not say where either cup
is, how wide to open the gripper, how far down to close it, or which pixels in
an image belong to the cup. Somebody has to supply those numbers, and they
depend on the object in front of the robot. Most language-driven robots have a
programmer supply them in advance: measure once, freeze the constants into
functions such as \texttt{find(object)} and \texttt{pick(object)}, and let the
model choose functions and fill in arguments
\cite{saycan2022, huang2022zeroshot, codeaspolicies2023, progprompt2023}. New
objects therefore need new programmer-supplied code and measurements. This
limits the flexibility and generalization capability of the system.

\begin{figure}[!t]
  \centering
  \includegraphics[width=\linewidth]{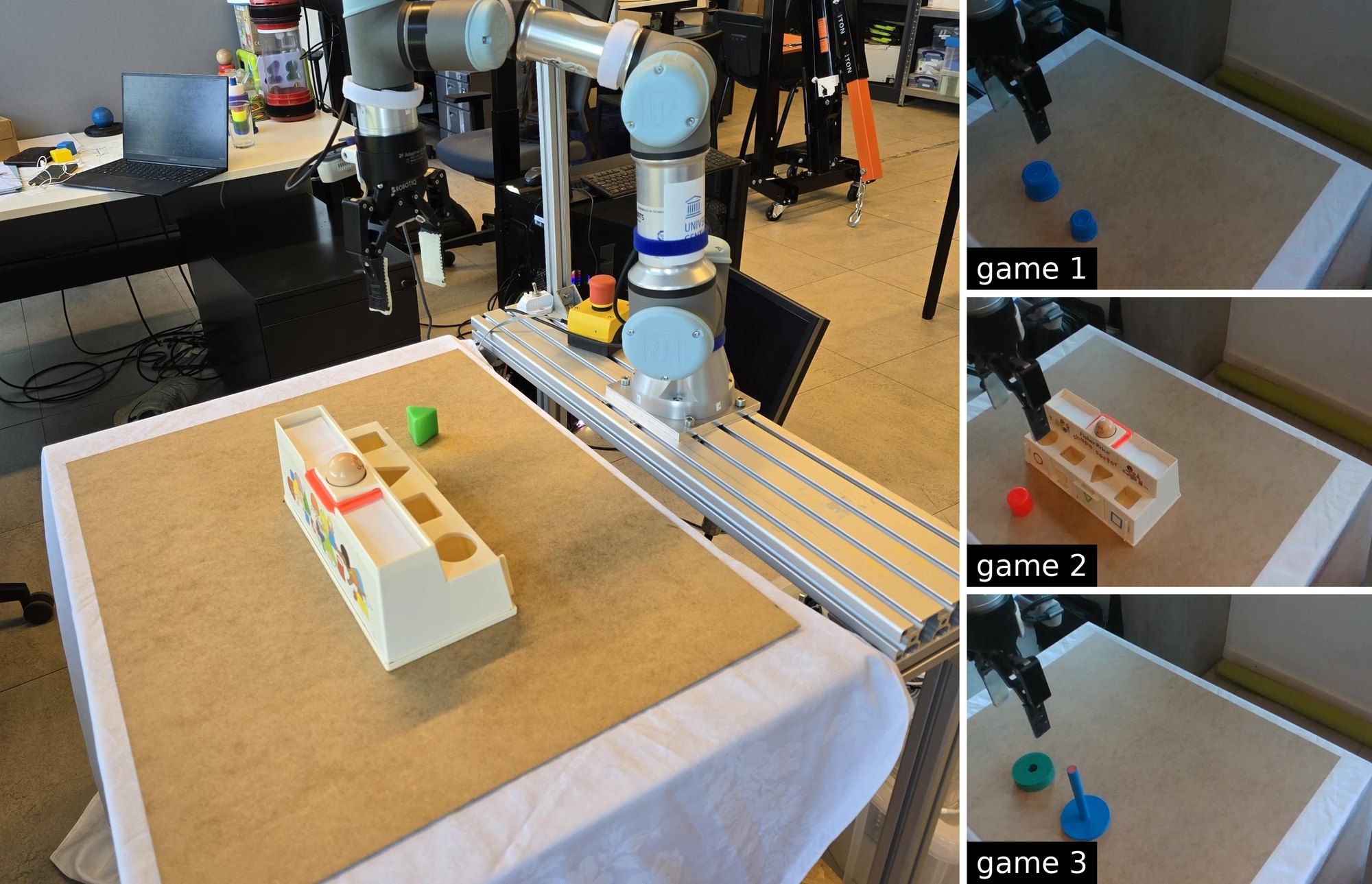}
  \caption{The robotic arm, here with the \textit{Sorter} game. Right: the basic tasks for which skills were provided (two blue cups, a red cylinder, the
  teal ring).}
  \label{fig:cell}
\end{figure}

Code as Policies \cite{codeaspolicies2023} had a large language model write
that code itself. However, the large language models of 2022 were weak
programmers. Furthermore, their system was open-loop, meaning the agent could
not debug its code. Both limits have since lifted. Open-weight models at
hundreds of billions of parameters (GLM~5.3\footnote{\url{https://huggingface.co/zai-org/GLM-5.3}.},
Kimi~K3\footnote{\url{https://github.com/MoonshotAI/Kimi-K3}.})
and even local models (Qwen3.8-27B\footnote{\label{fn:qwen}\url{https://huggingface.co/Qwen/Qwen3.8-27B}.}) now
write and debug code. Furthermore, agent harnesses give a model a shell, a
filesystem, a loop that hands tracebacks back to it
\cite{sweagent2024,swebench2024}, and load long procedural documents (skills)
only when they are needed.\footnote{\label{fn:skills}\url{https://www.anthropic.com/engineering/equipping-agents-for-the-real-world-with-agent-skills}.}

Recent work
~\cite{capx2026,alrm2026,aspire2026} puts coding agents on robots and
investigates their manipulation performance.
However, none of this work tests generalization capabilities on real-world hardware with a
locally running open-weight model.

In this work we investigate if a local model can generalize to new object-related variations of
a task. In our system, a locally served Qwen3.8-27B within a coding-agent harness
operates a UR3e arm with a parallel gripper and a wrist camera
(Fig.~\ref{fig:cell}). A service owns
the hardware and exposes nine low-level operations (Section~\ref{sec:method}). Its
task knowledge consists of two \emph{skills} per game: procedural documents,
each describing the task on one object instance.

We test our system on nine tabletop tasks built from children's toys designed
to probe generalization capability across object characteristics: color, size,
shape, and task variation. We further test if a speedup is present when
the agent is asked to complete the task a second time with a rearranged scene.

Supplementary materials, including a video of the trials, additional
experimental data, agent session logs, scene-camera recordings, the system
prompt, and skill documents, are available on the project
page.\footnote{\label{fn:supplementary}\url{https://rtalwar2.github.io/agentic-coding-for-robot-manipulation/}\\.
Video: \url{https://youtu.be/09Z0gJm21E4}.}

Our contributions are:
\begin{itemize}
  \item An investigation into the generalization capability and limitations of local
	  coding agents serving as manipulation controllers on physical hardware, including timings.
  \item A framework that lets a local coding agent control a robot by writing code
	or executing tool calls on top of a service implementing basic motion
	and computer vision primitives, starting from a skill. Our framework further
	enforces safety by limiting tool speed, gripper force and workspace bounds.
  \item A protocol for assessing generalization capability by using toys used in developmental psychology.
  \item Evidence that a coding agent improves across passes: resuming its own
      session on a rearranged scene halves the tool calls and cuts output tokens
      to a third.
\end{itemize}

%% file: sections/related.tex
\section{RELATED WORK}
\label{sec:background}

Language models have been used for a long time as controllers for robots.
The most common arrangement
asks the model for a plan over a fixed inventory of skills
\cite{huang2022zeroshot}: the model proposes named actions, an affordance
model scores their feasibility \cite{saycan2022}, and textual feedback from
the scene closes the loop \cite{innermonologue2022}. A second line asks for
a program against an API of perception and control functions
\cite{codeaspolicies2023,progprompt2023,voxposer2023}. This gains loops,
conditionals and arithmetic on coordinates, but the API functions are
task-level and perception is provided. Later work lets a model
write new low-level primitives from curated documentation
\cite{promptbook2024}, emit keypoint constraints per task instance with
perception from large vision models \cite{rekep2024}, or control real arms
zero-shot through code \cite{geminirobotics2025}. Constrained variants
guarantee the feasibility of the generated program \cite{proc3s2024}.
Frontier API-served models are now also benchmarked as direct robot
controllers.\footnote{Robocurve, ``GPT-6 Astra on robot arms,'' third-party benchmark: \url{https://openai.robocurve.org/gpt-6-astra/}.} End-to-end policies
\cite{rt2_2023,pi0_2024} avoid the vocabulary problem at the cost of a
demonstration corpus.

New developments in large language models and agentic coding led
to the development of harnesses. These give a model a shell, a
filesystem and an execution loop. Yang et al. \cite{sweagent2024} showed that a model can
iterate on a failing program, and scores on such benchmarks
\cite{swebench2024} rose sharply as models improved. Skill libraries that
grow inside such loops exist in simulation, as executable code with
self-verification \cite{voyager2024} and as reward code refined against a
simulator \cite{eureka2024}. Two recent preprints put coding agents on
manipulation. ALRM \cite{alrm2026} compares code-as-policy against
tool-as-policy across models in simulation. CaP-X \cite{capx2026}
benchmarks coding agents across interface-abstraction tiers on simulated and
real arms and finds that a higher-level interface helps small models most. ASPIRE \cite{aspire2026} has an API model discover text skills from its own failures in
simulation and reports with- and without-skill trials on a real bimanual arm. Outside robotics, SkillsBench \cite{skillsbench2026} reports that curated
procedural skills lift smaller models towards larger ones and that a few
focused skills outperform exhaustive bundles. 

One-shot and few-shot imitation conditions a policy on one or two demonstration
trajectories \cite{kat2024,instantpolicy2025}. Demo2Code compiles a
demonstration into code \cite{demo2code2023}, and TidyBot generalizes a handful
of user examples into preferences on physical hardware \cite{tidybot2023}.
Learned policies generalize to unseen objects and scenes when trained on
intermediate representations such as keypoints, measured over 2,000 real
rollouts \cite{lips2026kil}. Text has also carried task knowledge. DROC
\cite{droc2024} accumulates language corrections into a knowledge base for a
code-writing executor, and AutoManual \cite{automanual2024} has an agent
compile its own interaction into a rule manual that guides smaller models in
simulation. ELLMER \cite{ellmer2025} retrieves curated text for a model on a
real kitchen manipulator.

Building on these coding-agent approaches, we investigate generalization
across object and task variations within each game while holding the
model, low-level interface and skill documents fixed. Our agent uses a
locally deployed 27B vision-language model on a single workstation and
adapts procedures documented for one object instance without additional
human programming or model training. The agent writes additional perception and control code. 
We evaluate the capabilities and limitations of this configuration on physical hardware.

%% file: sections/method.tex
\section{METHOD}
\label{sec:method}

The system has three parts (Fig.~\ref{fig:stack}). The \emph{platform} owns the
robot and enforces its motion limits. It holds everything whose correctness
must not depend on the model. The \emph{agent setup} is the model, the system
prompt, the tools, the skills and a workspace template with the platform
manual, configured once per robot. In each \emph{session} the model reads,
writes and runs code in an isolated workspace and cannot modify the other two.
Invariant hardware functions (kinematics, limits, the geometry that turns a
pixel into a coordinate) live in the platform, and knowledge from one object
instance lives in a skill. Scene measurements such as object width should be
remeasured by the model in the session.

\input{figs/stack.tex}

\subsection{Platform layer}
\label{sec:method:platform}

The platform is one hardware-owning HTTP service that wraps airo-mono,\footnote{\url{https://github.com/airo-ugent/airo-mono}.}
a Python robotics library. TABLE~\ref{tab:sdk} lists nine
operations exposed by the platform. \texttt{locate\_pixel} maps one pixel to a point in base-frame coordinates using                                                                                                   
the depth map and the hand-eye calibration. \texttt{locate\_objects} is a                                                                                                    
color-blob detector with depth information. It takes its color
thresholds as required arguments and returns its evidence (sample counts, fit
residuals, warnings) with every answer. \texttt{descend\_to\_contact} exposes
the wrist force sensor. To avoid waiting for a model response at each control step, time-sensitive
feedback loops, such as stall detection and stepwise guarded descent,
run within the platform as blocking operations. There is no trained perception module. The platform                                                                                                           
supplies this one generic detector, while the model measures the                                                                                                        
color thresholds on the live frame and writes the rest of the perception                                                                                                          
during the run. The only learned component is the language model.

To enforce safety, every actuation request, from a tool or from code the model
wrote, passes through safety limits. Every Cartesian target is clamped into a
workspace box. Linear speed, joint speed and gripper force are capped, and the
service boots into a motion lock.

\begin{table}[!htb]
\centering
\small
\begin{tabular}{@{}lp{4.7cm}@{}}
\toprule
\textbf{Operation} & \textbf{Description} \\
\midrule
\texttt{look} & Capture a wrist-camera frame. \\
\texttt{get\_state} & Proprioception: TCP pose, joints, gripper width, the TCP force-torque reading, motion-lock state. \\
\texttt{move\_tcp} & Move the tool center point to a Cartesian pose. \\
\texttt{move\_joints} & Joint-space move. \\
\texttt{gripper} & Command width, force and speed. \\
\texttt{locate\_pixel} & Turn one image pixel into a base-frame point via depth and the hand-eye calibration. \\
\texttt{locate\_objects} & Fixed color-blob detector with depth. HSV bounds are required arguments. Returns center, top height, diameter and evidence per blob. \\
\texttt{descend\_to\_contact} & Step the tool down and stop on a force change, force read while stationary between steps. \\
\texttt{set\_motion} & Unlock or lock the motors. The service boots senses-only. \\
\bottomrule
\end{tabular}
\caption{The nine platform operations, available as harness tools and as
a Python SDK. Both reach the same service and the same safety limits.}
\label{tab:sdk}
\end{table}

\subsection{Agent setup}
\label{sec:method:setup}

We use \emph{pi},\footnote{\url{https://github.com/badlogic/pi-mono}.} a minimal open-source coding-agent
harness, and configure the model, the system prompt, the tools, the skills
and the workspace. 

The system prompt tells the model it operates
this robot, biases it towards measure-act-verify loops, and requires script
files rather than one-liners. Its central rule: constants of the workspace box
(camera offsets, table height, pad geometry) are to be trusted, while
numbers measured on an object are to be re-measured.

The tools are the nine operations of TABLE~\ref{tab:sdk} plus the
harness's file and shell tools. The model chooses to either execute a tool call
or chunk tool calls via scripting using the SDK.

The skills are hand-crafted procedural documents in natural language
(TABLE~\ref{tab:skills}). A skill holds the procedure step by step. Under each
step it lists the measured values that justify it (color thresholds, grasp
heights, contact forces) and how each was obtained, the verification steps, and
the pitfalls met during authoring, including checks that gave a false pass. The
values are marked as being valid only for the specific measured instance. The
procedure is the content meant to transfer. As in the harness convention,\footnotemark[\getrefnumber{fn:skills}]
only a skill's name and one-line description are in
the prompt. The body enters context when the task makes it relevant.
\begin{table}[!htb]
\centering
\footnotesize
\begin{tabular}{@{}p{1.75cm}rp{5.3cm}@{}}
\toprule
\textbf{Skill} & \textbf{Lines} & \textbf{Procedure} \\
\midrule
\texttt{grasp-\allowbreak a-\allowbreak cup} & 208 & Pick one cup: measure the color window live, locate by depth evidence, touch the top, grasp at the pad band with two closes, check the lift. \\
\texttt{stack-\allowbreak two-\allowbreak cups} & 192 & Stack one cup on another: locate both, grasp, find the release height by force, verify from a profile view. \\
\texttt{grasp-\allowbreak a-\allowbreak cylinder} & 187 & Pick a cylinder: read its orientation from the silhouette, center it with or without depth, check the stall width. \\
\texttt{insert-\allowbreak into-\allowbreak a-\allowbreak hole} & 324 & Insert into the round hole: identify the hole, measure it parallax-free, find the face by touch, use a force-free descent as the alignment test, verify by look. \\
\texttt{grasp-\allowbreak a-\allowbreak ring} & 182 & Pick a flat ring at the jaw's span limit: locate by color, straddle with a force-watched open-jaw descent, check the stall width. \\
\texttt{put-\allowbreak a-\allowbreak ring-\allowbreak on-\allowbreak a-\allowbreak rod} & 277 & Thread the held ring onto the rod: find the tip by its marker, predict the ground-out height, verify by a force-free descent and a top-down look. \\
\bottomrule
\end{tabular}
\caption{The six skills, one grasp and one placement per game.}
\label{tab:skills}
\end{table}

The workspace is a per-session directory copied fresh from a template,
so no state crosses between tasks. The template carries the platform manual
which consists of four pages describing the SDK, frames and conventions of the
platform layer, the camera's behavior, and the robot's measured constants such
as the maximum gripper width, minimum gripper width and home TCP pose.

%% file: figs/stack.tex
\begin{figure}[!htb]
\vspace*{5pt}%
\centering
\newcommand{\ex}[1]{{\scriptsize\color{black!60!black}#1}}%
\begin{tikzpicture}[
  font=\footnotesize,
  box/.style={draw, rounded corners=1pt, align=left, inner xsep=5pt, inner ysep=4pt, text width=#1},
  lab/.style={font=\bfseries\footnotesize},
  arr/.style={-{Stealth[length=4pt]}, line width=0.6pt},
  node distance=3mm and 3mm]
\node[box=3.4cm] (model) {\textbf{Model} Qwen3.8-27B, deployed locally\\[1pt]
  \scriptsize reasons, executes tool calls, writes and executes scripts};
\node[box=3.4cm, right=of model] (prompt) {\textbf{System prompt} 68 lines\\[1pt]
  \scriptsize tells the model it operates a robot, biases it towards measure-act-verify loops};
\node[box=2.15cm, below=7mm of model.south west, anchor=north west] (skills) {\textbf{Skills} 2 per game\\
  \ex{the procedure for one object: its measured values, how each was obtained, pitfalls while measuring}};
\node[box=2.15cm, right=of skills] (manual) {\textbf{Manual} 617 lines\\
  \ex{how this robot behaves for any object: gripper and camera geometry, table height, the SDK}};
\node[box=2.15cm, right=of manual] (script) {\textbf{Scripts} written this run\\
  \ex{measures the object in front of the arm, computes the grasp, acts, checks}};
\node[draw, dashed, gray, inner sep=3pt, fit=(skills)(manual)(script)] (ws) {};
\node[gray, font=\scriptsize, anchor=south east, inner sep=1pt] at (ws.north east) {workspace, fresh copy per session};
\node[box=3.4cm, below=2.5mm of ws.south west, anchor=north west] (tools) {\textbf{Tool call} one of 9 operations\\
  \ex{one action, its result straight into context}};
\node[box=3.4cm, below=2.5mm of ws.south east, anchor=north east] (sdk) {\textbf{SDK call} from the script\\
  \ex{Enables scripting of tool calls}};
\node[box=7.15cm, below=5mm of tools.south west, anchor=north west] (plat) {\textbf{Platform} one service owns the hardware; kinematics, workspace box, speed and force caps, motion lock;\\
	};
\node[box=7.15cm, below=3mm of plat] (hw) {\textbf{Robotic Arm} UR3e, Robotiq 2F-85, wrist RealSense D435, fixed table};
\draw[arr] (model) -- (prompt);
\draw[arr] (model.south) -- (ws.north -| model.south) node[midway, right, font=\scriptsize] {reads, writes};
\draw[arr] (model.west) -- ++(-2.5mm,0) |- (tools.west);
\draw[arr] (script.south) -- (sdk.north);
\draw[arr] (tools.south) -- (plat.north -| tools.south);
\draw[arr] (sdk.south) -- (plat.north -| sdk.south);
\draw[arr] (plat.south) -- (hw.north);
\node[font=\scriptsize, anchor=north] at ($(tools.south)!0.5!(sdk.south) + (0,-1.8mm)$) {same endpoints and safety limits};
\end{tikzpicture}
\caption{Architectural overview of the framework and the interacting components.}
\label{fig:stack}
\end{figure}
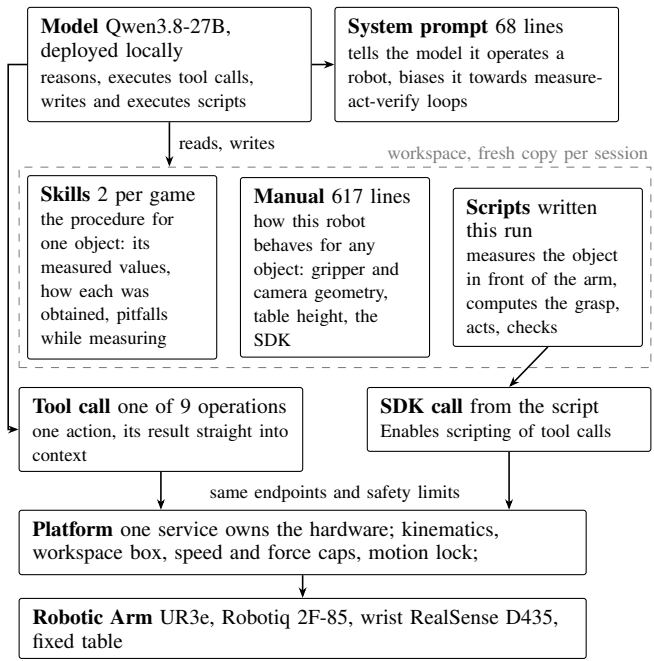

%% file: sections/experiments.tex
\section{EXPERIMENTAL DESIGN}
\label{sec:experiments}

We investigate whether a local open-weight vision-language model can control a
robot and one-shot generalize to new variations of a task without new human
programming or training on physical hardware. Specifically, we test it on nine
tasks built from children's toys designed to probe generalization capability
across various object characteristics: color, scale, shape, and simple
task variations. We measure whether it can accomplish unseen object
variations of a task, how long it takes to complete those tasks and investigate
the limitations of such a system.

\subsection{Games and Tasks}
\label{sec:experiments:design}

In order to probe generalization capacity along object variations and task
variations, we employ children's toys. They are particularly well suited as they
give compact, measurable families of related tasks: the objective stays
recognizable while color, scale, shape or task changes. For the same reason,
developmental psychology uses them as graded instruments of skill transfer
\cite{ornkloo2007,greenfield1972}. We only investigate object-related and task-related
generalizations. We do not investigate generalization across other
variations such as material, lighting, clutter, workspace layout, task
vocabulary and robot morphology. The three games we employ are: \textit{Cups},
a \textit{Sorter}, and \textit{Rings}
(Fig.~\ref{fig:objects}). We provide for each game, as discussed in
Section~\ref{sec:method:setup}, two skills describing a single task
from which we then test variations. With each game we perform several tasks,
listed in TABLE~\ref{tab:axes} with their prompts, generalization axes and
time caps.

\begin{table*}[!t]
\vspace*{5pt}%
\centering
\footnotesize
\begin{tabular}{@{}p{3.1cm}@{\hspace{1em}}p{5.0cm}@{\hspace{1em}}p{2.8cm}@{\hspace{1em}}cccc@{\hspace{1em}}c@{}}
\toprule
& & & \multicolumn{4}{c}{\textbf{Variation Axes}} & \\
\cmidrule(lr){4-7}
\textbf{Tasks} & \textbf{Prompt} & \textbf{On the table} & \textbf{Color} & \textbf{Size} & \textbf{Shape} & \textbf{Task Var.} & \textbf{Time Cap} \\
\midrule
Cups: Red Cups & Stack the smaller red cup on top of the larger red cup. & medium red, large red & $\bullet$ & $\bullet$ & & & 22\,min \\
\addlinespace
Cups: Large blue cups & Stack the small blue cup on top of the large blue cup. & small blue, large blue & & $\bullet$ & & & 22\,min \\
\addlinespace
Cups: Yellow and Green Cups & Stack the yellow cup on top of the green cup. & small yellow, large green & $\bullet$ & $\bullet$ & & & 22\,min \\
\addlinespace
Cups: Three Cups & Stack the three cups on the table into one tower, largest at the bottom. & large yellow, medium green, small green & $\bullet$ & $\bullet$ & & $\bullet$ & 66\,min \\
\addlinespace
Cups: Sort Cups & Arrange the four cups on the table in a straight line, from smallest to largest. & large green, medium yellow, small blue, small red & $\bullet$ & $\bullet$ & & $\bullet$ & 88\,min \\
\addlinespace
Sorter: Beam & Put the yellow rectangular block into its matching hole in the shape sorter. & sorter, yellow beam & $\bullet$ & & $\bullet$ & & 68\,min \\
\addlinespace
Sorter: Prism & Put the green triangle block into its matching hole in the shape sorter. & sorter, green triangle & $\bullet$ & & $\bullet$ & & 68\,min \\
\addlinespace
Sorter: Cube & Put the blue cube into its matching hole in the shape sorter. & sorter, blue cube & $\bullet$ & & $\bullet$ & & 68\,min \\
\addlinespace
Rings & Stack all the rings on the table onto the rod to rebuild the tower, largest at the bottom. & rod, base, all five rings & $\bullet$ & $\bullet$ & & $\bullet$ & 316\,min \\
\bottomrule
\end{tabular}
\caption{The four generalization axes from the base task for each new task, with each
task's prompt and time cap. A bullet marks an axis the task varies relative
to the instance its skill documents.}
\label{tab:axes}
\end{table*}

\input{figs/objects.tex}

The \textit{Cups} game uses twelve nesting cups with rim diameters
ranging from 36 to 85\,mm, in four repeating colors and stacked upside down.
With this game we probe generalization across color, size and task variation.
The basic task documented for this game in a skill is the single stacking of a
small blue cup on top of a larger blue cup. From this setup we test five
variations. The first asks of the system to do the task with red cups instead.
The second asks to repeat the task with different-sized blue cups. The third
combines the two former and asks the system to stack a small yellow cup on top
of a larger green cup. The fourth combines scale and color and introduces task
variation, asking to stack three cups with varying colors and size. The fifth
is a completely different task: we ask the system to order four cups in a
straight line from smallest to largest.

The second game, \textit{Sorter}, is a peg-in-hole game. It consists of a
base with four different holes varying in shape and accompanying pegs which
also vary in color. In order, there is a circular hole with a red cylinder, a
rectangular hole with a yellow beam, a triangular hole with a green prism and a
square hole with a blue cube. The pegs fit their holes with about 1\,mm of
clearance per side. With this game we probe shape generalization. The skill we
provide for this game describes insertion of the red cylinder in the circular
hole. We then ask the system to insert the other pegs as well. The difficulty
within this game is ordered by yaw symmetry: the cylinder (the skill's object)
fits at any wrist angle, the cube every 90$^\circ$, the triangle every
120$^\circ$, the beam every 180$^\circ$.

The third game, \textit{Rings}, consists of five rings and a rod.
The five rings vary in size, ranging from 40 to 80\,mm, and color. The rings
fit on the rod with a clearance of 2\,mm. With this game we ask the system
to perform a single task: stack the rings in order on the rod. This task combines
color, size and task variation. The skill we provided upfront documents how
to grab the largest, teal ring and insert it on the rod.

For each task we define a time cap upfront so no experiment can last
indefinitely. This time cap is derived from the basic tasks documented in the
skills multiplied by four to account for the added complexity from
generalization, and again multiplied by the number of objects the robot has to
pick up.

\subsection{Setup}
\label{sec:experiments:setup}

The robotic arm is a UR3e arm on a fixed table equipped with a Robotiq 2F-85
parallel gripper with the tool center point at the fingertip pinch, and an
Intel RealSense D435 color-depth camera on the wrist, eye-in-hand calibrated.

The model we employ is Qwen3.8-27B,\footnotemark[\getrefnumber{fn:qwen}] a dense 27B vision-language
model, served with vLLM 0.27.1 (FP8 weights and KV cache, speculative decoding
with three draft tokens) on the workstation that also hosts the platform
(NVIDIA RTX PRO 6000, 96\,GB). The context window is 262{,}144 tokens.
Sampling uses the model's default inference parameters (temperature 1.0, top-$p$ 0.95, top-$k$ 20) and reasoning runs at effort \emph{low}.
Reasoning is the process of generating thinking
steps before producing a final answer.

\subsection{Scoring and interventions}

We perform five trials for each task, from now on abbreviated as
\emph{t1}--\emph{t5}, and record the following quantities: the
success or failure, the duration, the tokens used, the number of tool calls and
the percentage of tokens dedicated to reasoning. In the ideal case the agent
completes the task without problems. However, sometimes an intervention is
necessary. We distinguish between three interventions: small intervention,
emergency stop and protective stop. A small intervention happens when a mistake
by the agent results in an object lying sideways or when an object falls from
the table. When that happens, we put the object upright and back on the table,
respectively. Furthermore, a protective stop or emergency stop intervention is
allowed for safety reasons. A protective stop is triggered by the robotic arm
while the emergency stop is triggered by the operator.

\subsection{Second Pass}

We further test if there is a speedup when the agent is asked to do a second
pass after successful completion of the task. In order to evaluate this we
resume the sessions of the fastest and slowest successful trials of each task and
ask the agent to perform the task again. The prompt we use for this is the
following:
\begin{quote}\small
``That worked - the task was completed correctly. The scene has been reset
and the objects are in new positions. Do the task again: \emph{<original task
prompt>}''
\end{quote}
We measure the same quantities as in the first passes of each task in order to
determine if there is a possible improvement.

%% file: figs/objects.tex
\begin{figure}[!htb]
\centering
\begin{tikzpicture}[line width=0.4pt]
\footnotesize
\node[anchor=west,font=\footnotesize\itshape] at (0,1.10) {Cups, rim 36--85\,mm};
\draw[fill=blue!70,line width=0.4pt] (0.000,0) -- (0.714,0) -- (0.614,0.546) -- (0.100,0.546) -- cycle;
\node[font=\tiny,anchor=north] at (0.357,-0.03) {85\,mm};
\draw[fill=green!60!black!70,line width=0.4pt] (0.904,0) -- (1.584,0) -- (1.489,0.530) -- (1.000,0.530) -- cycle;
\node[font=\tiny,anchor=north] at (1.244,-0.03) {81\,mm};
\draw[fill=yellow!85!orange,line width=0.4pt] (1.774,0) -- (2.405,0) -- (2.316,0.504) -- (1.862,0.504) -- cycle;
\node[font=\tiny,anchor=north] at (2.090,-0.03) {75\,mm};
\draw[fill=red!75,line width=0.4pt] (2.595,0) -- (3.191,0) -- (3.108,0.479) -- (2.678,0.479) -- cycle;
\node[font=\tiny,anchor=north] at (2.893,-0.03) {71\,mm};
\draw[fill=blue!70,line width=1.4pt] (3.381,0) -- (3.935,0) -- (3.858,0.454) -- (3.458,0.454) -- cycle;
\node[font=\tiny,anchor=north] at (3.658,-0.03) {66\,mm};
\draw[fill=green!60!black!70,line width=0.4pt] (4.125,0) -- (4.646,0) -- (4.573,0.437) -- (4.198,0.437) -- cycle;
\node[font=\tiny,anchor=north] at (4.386,-0.03) {62\,mm};
\draw[fill=yellow!85!orange,line width=0.4pt] (4.836,0) -- (5.314,0) -- (5.248,0.403) -- (4.903,0.403) -- cycle;
\node[font=\tiny,anchor=north] at (5.075,-0.03) {57\,mm};
\draw[fill=red!75,line width=0.4pt] (5.504,0) -- (5.949,0) -- (5.887,0.378) -- (5.566,0.378) -- cycle;
\node[font=\tiny,anchor=north] at (5.727,-0.03) {53\,mm};
\draw[fill=blue!70,line width=1.4pt] (6.139,0) -- (6.542,0) -- (6.486,0.353) -- (6.196,0.353) -- cycle;
\node[font=\tiny,anchor=north] at (6.341,-0.03) {48\,mm};
\draw[fill=green!60!black!70,line width=0.4pt] (6.732,0) -- (7.102,0) -- (7.050,0.328) -- (6.784,0.328) -- cycle;
\node[font=\tiny,anchor=north] at (6.917,-0.03) {44\,mm};
\draw[fill=yellow!85!orange,line width=0.4pt] (7.292,0) -- (7.628,0) -- (7.581,0.302) -- (7.339,0.302) -- cycle;
\node[font=\tiny,anchor=north] at (7.460,-0.03) {40\,mm};
\draw[fill=red!75,line width=0.4pt] (7.818,0) -- (8.120,0) -- (8.078,0.252) -- (7.860,0.252) -- cycle;
\node[font=\tiny,anchor=north] at (7.969,-0.03) {36\,mm};
\node[anchor=west,font=\footnotesize\itshape] at (0,-0.98) {Sorter, pegs 28--45\,mm};
\draw[fill=gray!10,line width=0.4pt,rounded corners=2pt] (1.470,-2.10) rectangle (4.311,-1.26);
\draw[line width=0.4pt] (1.884,-1.680) circle (0.225);
\draw[line width=0.4pt] (2.285,-1.839) rectangle (2.774,-1.521);
\draw[line width=0.4pt] (2.950,-1.874) -- (3.401,-1.874) -- (3.176,-1.486) -- cycle;
\draw[line width=0.4pt] (3.689,-1.869) rectangle (4.067,-1.491);
\node[font=\tiny,anchor=north] at (2.890,-2.13) {base 290$\times$140\,mm};
\draw[fill=red!75,line width=1.4pt] (0.400,-2.950) circle (0.215);
\node[font=\tiny,anchor=north,align=center] at (0.400,-3.290) {cylinder 41\,mm\\any yaw};
\draw[fill=blue!70,line width=0.4pt] (1.902,-3.129) rectangle (2.259,-2.771);
\node[font=\tiny,anchor=north,align=center] at (2.081,-3.290) {cube 34\,mm\\90$^\circ$};
\draw[fill=green!60!black!70,line width=0.4pt] (3.472,-3.136) -- (3.903,-3.136) -- (3.688,-2.764) -- cycle;
\node[font=\tiny,anchor=north,align=center] at (3.688,-3.290) {triangle 41\,mm\\120$^\circ$};
\draw[fill=yellow!85!orange,line width=0.4pt] (5.134,-3.099) rectangle (5.602,-2.801);
\node[font=\tiny,anchor=north,align=center] at (5.368,-3.290) {beam 44.6$\times$28.3\,mm\\180$^\circ$};
\node[anchor=west,font=\footnotesize\itshape] at (0,-4.050) {Rings, outer 40--80\,mm};
\draw[fill=green!60!black!70,line width=1.4pt,even odd rule] (0.450,-4.970) circle (0.420) (0.450,-4.970) circle (0.110);
\node[font=\tiny,anchor=north] at (0.450,-5.420) {80\,mm};
\draw[fill=violet!70,line width=0.4pt,even odd rule] (1.840,-4.970) circle (0.368) (1.840,-4.970) circle (0.110);
\node[font=\tiny,anchor=north] at (1.840,-5.367) {70\,mm};
\draw[fill=yellow!80,line width=0.4pt,even odd rule] (3.125,-4.970) circle (0.315) (3.125,-4.970) circle (0.110);
\node[font=\tiny,anchor=north] at (3.125,-5.315) {60\,mm};
\draw[fill=gray!15,line width=0.4pt,even odd rule] (4.305,-4.970) circle (0.263) (4.305,-4.970) circle (0.110);
\node[font=\tiny,anchor=north] at (4.305,-5.262) {50\,mm};
\draw[fill=pink!80,line width=0.4pt,even odd rule] (5.380,-4.970) circle (0.210) (5.380,-4.970) circle (0.110);
\node[font=\tiny,anchor=north] at (5.380,-5.210) {40\,mm};
\draw[fill=blue!70,line width=0.4pt] (6.40,-5.390) rectangle (6.60,-4.550);
\draw[fill=blue!70!black!40,line width=0.4pt] (6.05,-5.470) rectangle (6.95,-5.390);
\node[font=\tiny,anchor=north,align=center] at (6.50,-5.510) {rod height 155\,mm};
\end{tikzpicture}
\caption{Illustration of the objects involved in the games. Heavy outline indicates the
object was included in a skill.}
\label{fig:objects}
\end{figure}
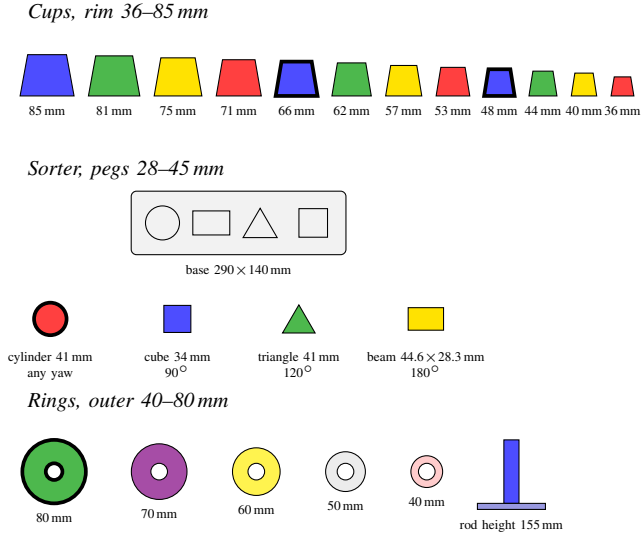

%% file: sections/discussion.tex
\section{RESULTS}
\label{sec:results}

We investigated if a local open-weight vision-language model can control a
robotic arm and generalize to new variations of a task. In order to evaluate
this, we tested such a model, Qwen3.8-27B, on three children's games capable of
probing generalization across various axes (TABLE~\ref{tab:axes}). We derived
nine tasks from the three base tasks documented in the skills, and ran each
five times. TABLE~\ref{tab:grid} reports the per-task results.

\begin{table*}[!t]
  \vspace*{5pt}%
  \centering
  \footnotesize
  \begin{tabular}{@{}l@{\hspace{0.8em}}c@{\hspace{0.8em}}rrr@{\hspace{0.8em}}c@{\hspace{0.8em}}cc@{\hspace{0.8em}}ccc@{
  }}
  \toprule
  & & \multicolumn{3}{c}{\textbf{Duration} \emph{(min)}} & & & & \multicolumn{3}{c}{\textbf{Interventions} \emph{(\#)}}
  \\
  \cmidrule(lr){3-5}\cmidrule(lr){9-11}
  \textbf{Task} & \parbox[b]{1.7cm}{\centering\textbf{Success\\ (no interv.)}} & \textbf{Min} & \textbf{Med} &
  \textbf{Max} & \textbf{Timeouts} \emph{(\#)} & \textbf{Tokens (Median)} & \textbf{Reas.} \emph{(\%)} & \textbf{Small}
  & \textbf{E-stop} & \textbf{Prot.\ stop} \\
  \midrule
  Cups: Red Cups & 4/5 & 3.8 & 4.4 & 22.0 & 1 & 14.7k & 71 & 1 & -- & -- \\
  Cups: Large blue cups & 5/5 & 3.4 & 6.1 & 6.7 & -- & 23.5k & 76 & -- & -- & -- \\
  Cups: Yellow and Green Cups & 5/5 & 3.8 & 7.2 & 9.9 & -- & 30.8k & 74 & -- & -- & -- \\
  Cups: Three Cups & 3/5 & 8.0 & 12.7 & 14.6 & -- & 40.5k & 77 & 1 & -- & 1 \\
  Cups: Sort Cups & 3/5 & 12.0 & 13.8 & 15.8 & -- & 43.3k & 80 & 1 & -- & 1 \\
  \addlinespace
  Sorter: Cube & 4/5 & 12.9 & 28.8 & 60.8 & -- & 105.1k & 82 & -- & -- & 1 \\
  Sorter: Prism & 1/5 & 11.8 & 32.4 & 67.2 & 1 & 111.2k & 85 & 1 & 1 & 2 \\
  Sorter: Beam & 2/5 & 38.4 & 42.6 & 67.5 & -- & 150.2k & 82 & 2 & -- & 1 \\
  \addlinespace
  Rings & 3/5 & 22.8 & 29.8 & 59.0 & -- & 99.1k & 73 & -- & -- & 1 \\
  \bottomrule
  \end{tabular}
  \caption{Outcomes of the nine tasks at $N{=}5$. \emph{Success} counts only trials
  that finished without any operator intervention. Durations are wall-clock minutes. \emph{Timeouts} are trials stopped at the task's time cap.
  \emph{Tokens} is the median output tokens per trial and \emph{Reas.} the median
  share of those spent on reasoning. Interventions are counted per category as
  defined in \secref{sec:experiments}. Fig.~\ref{fig:durations} plots the
  durations.}
  \label{tab:grid}
\end{table*}

\subsection{General Behavior}

In general the grid took 15.8 robot-hours, produced 3.18 million output tokens,
resulted in 2,043 tool calls and executed 1,034 motion commands. Out of all 45
trials 30 were successful without any intervention. Across all trials the agent
followed a similar pattern. It first read the skill and relevant manual pages.
Then it measured the scene using self-authored perception scripts and proceeded
with writing pick-and-place scripts and executing them. Finally, it verified
what it did by looking from a predefined pose. In case of an error, it looped
back to measuring the scene. Reasoning accounted for 71\% to 85\% of output tokens in the task-level
medians, with reasoning share increasing with task complexity
 (\textit{Sorter}), therefore increasing task duration.
This is partially caused by our identity file which biased the model
toward chunking actions, and partially by enabling reasoning within the model.
Empirically we observed that this mode of operation caused a lower
sense-think-act frequency and more deliberate action, resulting in a higher success
rate and faster task completion. The longest thinking turns happened following
a surprise such as a jam or a contradiction between measurements. Trials
containing them took longer in general and resulted in outlier trial durations.
This occurred most in game two and game three. Out of all tasks, 30 ended
without any intervention, six required a small intervention, one an emergency
stop and we encountered seven protective stops.

The biggest successes can be attributed to two things. First, the skills
document tasks as formulas and procedures with certain values. The agent
successfully used these and re-measured values when necessary. Second,
verification behavior caught failures when they happened. For example, a stall
width that did not hold after the lift proved a grasp failed. The agent then
knew to try again.

\subsection{Failure Modes}

We encountered three failure modes: perception failures, anchoring on faulty
assumptions, and unawareness of the robot's morphology and the direct space around
it.

The first category of failures consists of perception failures, a wrong reading of the
scene. In \textit{Sorter: Cube} t4 the agent named the rectangular opening the square one and
kept attempting to put it in this hole with increasing force, until the arm triggered a protective stop. In \textit{Sorter: Prism} t1 and \textit{Sorter: Prism} t3
it read the standing prism as lying on its side and attempted to grasp it from
an incorrect approach angle leading to an emergency stop to prevent
self-collision. In \textit{Rings} t1 it measured the pink ring larger than the white one
from an oblique view and built the tower in the wrong order. In \textit{Rings} t2 the
white-ring detector returned the yellow ring and the arm descended onto the
rod, pressing the rod into the table, leading to a protective stop.
We believe this category of errors can be attributed to the limitations of
classical computer vision and can possibly be resolved by using more advanced
deep learning methods.

The second category consists of failures where the agent anchors itself on a faulty
assumption. Although the agent recovered from a faulty assumption in 10 out of
30 successful trials, in four runs it did not, leading to a failure. In \textit{Sorter: Cube} t4 it
measured the clearance nine different ways but never asked whether it had the
right hole. In \textit{Sorter: Prism} t2 it swept either the wrist angle or the position for forty
minutes but never swept them together. This led to either the position being
correct but not the angle or the angle being correct but not the position. In
\textit{Cups: Red Cups} t5 the agent performed a measurement from a pose from which it couldn't be
made. It realized the measurement was wrong but proceeded to remeasure from the
faulty pose. In \textit{Sorter: Beam} t4 it thought the square hole was the rectangular hole
and never realized this was wrong. This led to an outlier duration and ended
with the agent descending into the sorter and triggering a protective stop.
We think this category of errors is mostly intrinsic to the model we used.

The third category consists of failures caused by an unawareness of morphology and
the direct space around it. The agent has no body schema, and the platform's
checks cover only the tool center point. In \textit{Sorter: Prism} t1 and \textit{Sorter: Prism} t3 it rotated the tool
ninety degrees in mid-air. The point stayed inside the workspace box while the
gripper body swept toward the arm in one case and into the table in the other.
This led to an emergency stop and a protective stop, respectively. In \textit{Cups: Three Cups} t5 the agent
descended onto a cup with the jaws closed instead of open, in \textit{Sorter: Beam} t4 and \textit{Sorter: Prism} t4
onto the sorter with the jaws open, and in \textit{Cups: Sort Cups} t5 a fingertip landed on the cup
it had placed a moment earlier. In all cases, the agent pressed an object into
the table leading to a protective stop. We think this can be resolved by handing
the agent a simulation environment which includes its own morphology and
possibly the scene. A different solution could be including a scene camera
so the agent can see itself.

The task the agent did worst on is the Sorter: Prism. There was no single
leading cause but rather a combination of all the above categories.

\subsection{Second Pass Results}
\label{sec:experiments:secondpass}

\input{figs/timeline.tex}

A second pass on an already solved task was faster and cheaper across the board
(TABLE~\ref{tab:secondpass}, Fig.~\ref{fig:durations}). Over the 17 resumed
trials, the second pass completed the task two times faster (time ratio 0.48). The
agent used about half the tool calls of the first pass (354 against
734, ratio 0.48) and about a third of the output tokens (ratio 0.34). The output tokens decreased proportionally more 
than tool calls, while the median reasoning share of output tokens fell from 80\% to 67\%.
This is consistent with reduced reasoning demands when reusing prior work.
The agent kept the conclusions from the first pass, such as the color 
windows, the object heights, the identity of the matching hole and the wrist
angle that worked, and spent its remaining effort on re-locating the objects in
the new layout. For the cups this brought every task from four to fourteen
minutes down to two to eight minutes. For the sorter and the rings the gains
were larger in absolute terms but less uniform, and the second pass did not
remove the first-pass failure modes: one resumed beam trial ended in a
protective stop, and one resumed ring trial took longer than its first pass
because the agent failed to grasp the first ring five times before the rest
went quickly. The resumed trial's rank transferred. In six of the eight tasks
with two second passes, the resume of the faster first-pass trial was again the
faster one. What carries over is therefore not only the set of measured facts
but also the quality of the plan the first pass arrived at: a first pass that
found a clean path re-executes it cleanly, while one that reached its result by
a detour tends to repeat part of the detour. There were two exceptions where
the second pass hit a failure mode and took longer
(Fig.~\ref{fig:durations}). Sixteen of the seventeen second passes completed
without intervention.

These gains should be interpreted as in session reuse rather than be seen as
evidence of persistent learning. Resumed sessions retain conversation
history, measurements and generated scripts, and our experiment does not
isolate their individual contributions to the speedup. Nevertheless, the
improvement on rearranged scenes is promising: the agent can reuse prior
work without repeating the full discovery process. This motivates future
experiments on whether that experience can be consolidated into skills
that benefit fresh sessions.

\begin{table}[!htb]
\centering
\scriptsize
\begin{tabular*}{\columnwidth}{@{}l@{\extracolsep{\fill}}r@{\hspace{0.5em}}r@{\hspace{0.5em}}c@{\hspace{0.8em}}r@{\hspace{0.5em}}r@{\hspace{0.5em}}c@{}}
\toprule
& \multicolumn{3}{c}{\textbf{Tool calls}} & \multicolumn{3}{c}{\textbf{Tokens}} \\
\cmidrule(lr){2-4}\cmidrule(lr){5-7}
\textbf{Trial} & \textbf{1st} & \textbf{2nd} & \textbf{Ratio} & \textbf{1st} & \textbf{2nd} & \textbf{Ratio} \\
\midrule
Cups: Red Cups t2 & 24 & 12 & 0.50 & 23.5k & 6.0k & 0.26 \\
Cups: Red Cups t3 & 18 & 9 & 0.50 & 12.6k & 4.0k & 0.32 \\
Cups: Large blue cups t3 & 11 & 4 & 0.36 & 13.0k & 3.5k & 0.27 \\
Cups: Large blue cups t5 & 23 & 13 & 0.57 & 28.6k & 9.1k & 0.32 \\
Cups: Yellow and Green Cups t2 & 16 & 8 & 0.50 & 13.8k & 3.6k & 0.26 \\
Cups: Yellow and Green Cups t5 & 24 & 14 & 0.58 & 42.5k & 7.3k & 0.17 \\
Cups: Three Cups t1 & 22 & 14 & 0.64 & 27.9k & 11.4k & 0.41 \\
Cups: Three Cups t3 & 30 & 18 & 0.60 & 59.7k & 13.8k & 0.23 \\
Cups: Sort Cups t2 & 24 & 9 & 0.38 & 39.8k & 9.9k & 0.25 \\
Cups: Sort Cups t3 & 31 & 17 & 0.55 & 59.7k & 15.2k & 0.26 \\
\addlinespace
Sorter: Cube t3 & 105 & 32 & 0.30 & 145.2k & 25.7k & 0.18 \\
Sorter: Cube t5 & 29 & 15 & 0.52 & 57.2k & 23.5k & 0.41 \\
Sorter: Prism t5 & 85 & 32 & 0.38 & 111.2k & 21.4k & 0.19 \\
Sorter: Beam t1 & 88 & 22 & 0.25 & 108.2k & 25.0k & 0.23 \\
Sorter: Beam t5 & 70 & 43 & 0.61 & 109.3k & 58.3k & 0.53 \\
\addlinespace
Rings t3 & 51 & 55 & 1.08 & 83.8k & 86.6k & 1.03 \\
Rings t5 & 83 & 37 & 0.45 & 118.0k & 29.6k & 0.25 \\
\midrule
\textbf{All 17} & \textbf{734} & \textbf{354} & \textbf{0.48} & \textbf{1054.1k} & \textbf{354.0k} & \textbf{0.34} \\
\bottomrule
\end{tabular*}
\caption{The 17 second passes, each a resumed (fastest/slowest) first-pass trial of the task. Tool
  calls and output tokens for the first and second pass. \emph{Ratio} is the
  second-pass value over the first-pass value. Totals over all 17 in the last row.}
\label{tab:secondpass}
\end{table}

%% file: figs/timeline.tex
\newlength{\pw}
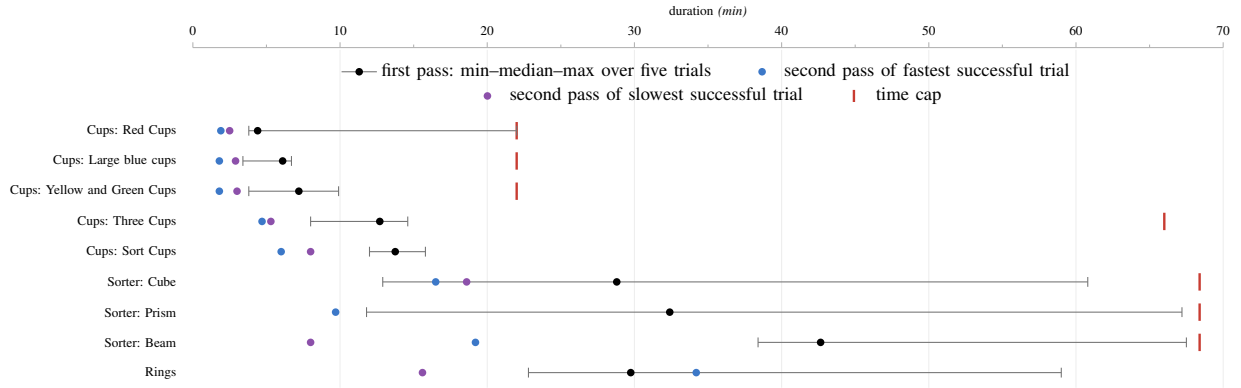
\begin{figure*}[!htb]
  \vspace*{2pt}%
  \centering
  \footnotesize
  \definecolor{p2blue}{RGB}{60,120,200}
  \definecolor{p2purple}{RGB}{140,80,170}
  \definecolor{capred}{RGB}{200,60,50}
  \setlength{\pw}{\dimexpr\textwidth-4.15cm\relax}
  \begin{tikzpicture}[line width=0.4pt,y=1cm]
  \draw[black!30] (0,0.62)--(1.00000*\pw,0.62);
  \draw[black!30] (0.00000*\pw,0.62)--(0.00000*\pw,0.690);
  \node[font=\tiny,anchor=south] at (0.00000*\pw,0.68) {0};
  \draw[black!30] (0.07143*\pw,0.62)--(0.07143*\pw,0.660);
  \draw[black!30] (0.14286*\pw,0.62)--(0.14286*\pw,0.690);
  \node[font=\tiny,anchor=south] at (0.14286*\pw,0.68) {10};
  \draw[black!30] (0.21429*\pw,0.62)--(0.21429*\pw,0.660);
  \draw[black!30] (0.28571*\pw,0.62)--(0.28571*\pw,0.690);
  \node[font=\tiny,anchor=south] at (0.28571*\pw,0.68) {20};
  \draw[black!30] (0.35714*\pw,0.62)--(0.35714*\pw,0.660);
  \draw[black!30] (0.42857*\pw,0.62)--(0.42857*\pw,0.690);
  \node[font=\tiny,anchor=south] at (0.42857*\pw,0.68) {30};
  \draw[black!30] (0.50000*\pw,0.62)--(0.50000*\pw,0.660);
  \draw[black!30] (0.57143*\pw,0.62)--(0.57143*\pw,0.690);
  \node[font=\tiny,anchor=south] at (0.57143*\pw,0.68) {40};
  \draw[black!30] (0.64286*\pw,0.62)--(0.64286*\pw,0.660);
  \draw[black!30] (0.71429*\pw,0.62)--(0.71429*\pw,0.690);
  \node[font=\tiny,anchor=south] at (0.71429*\pw,0.68) {50};
  \draw[black!30] (0.78571*\pw,0.62)--(0.78571*\pw,0.660);
  \draw[black!30] (0.85714*\pw,0.62)--(0.85714*\pw,0.690);
  \node[font=\tiny,anchor=south] at (0.85714*\pw,0.68) {60};
  \draw[black!30] (0.92857*\pw,0.62)--(0.92857*\pw,0.660);
  \draw[black!30] (1.00000*\pw,0.62)--(1.00000*\pw,0.690);
  \node[font=\tiny,anchor=south] at (1.00000*\pw,0.68) {70};
  \node[font=\tiny,anchor=south] at (0.50000*\pw,0.92) {duration \emph{(min)}};
  \draw[black!7,line width=0.3pt] (0.14286*\pw,0.59)--(0.14286*\pw,-3.880);
  \draw[black!7,line width=0.3pt] (0.28571*\pw,0.59)--(0.28571*\pw,-3.880);
  \draw[black!7,line width=0.3pt] (0.42857*\pw,0.59)--(0.42857*\pw,-3.880);
  \draw[black!7,line width=0.3pt] (0.57143*\pw,0.59)--(0.57143*\pw,-3.880);
  \draw[black!7,line width=0.3pt] (0.71429*\pw,0.59)--(0.71429*\pw,-3.880);
  \draw[black!7,line width=0.3pt] (0.85714*\pw,0.59)--(0.85714*\pw,-3.880);
  \draw[black!7,line width=0.3pt] (1.00000*\pw,0.59)--(1.00000*\pw,-3.880);
  \draw[black!55] (0.14432*\pw,0.30)--(0.17860*\pw,0.30);
  \fill (0.16146*\pw,0.30) circle (0.05);
  \node[font=\scriptsize,anchor=west] at (0.17564*\pw,0.30) {first pass: min--median--max over five trials};
  \fill[p2blue] (0.55249*\pw,0.30) circle (0.05);
  \node[font=\scriptsize,anchor=west] at (0.56667*\pw,0.30) {second pass of fastest successful trial};
  \fill[p2purple] (0.28591*\pw,-0.02) circle (0.05);
  \node[font=\scriptsize,anchor=west] at (0.30009*\pw,-0.02) {second pass of slowest successful trial};
  \draw[capred,line width=0.9pt] (0.64161*\pw,-0.10)--(0.64161*\pw,0.06);
  \node[font=\scriptsize,anchor=west] at (0.65579*\pw,-0.02) {time cap};
  \begin{scope}[yshift=-0.48cm]
  \node[font=\tiny,anchor=east] at (-0.12,0.000) {Cups: Red Cups};
  \draw[capred,line width=0.9pt] (0.31429*\pw,-0.115)--(0.31429*\pw,0.115);
  \draw[black!55] (0.05429*\pw,0.000)--(0.31429*\pw,0.000);
  \draw[black!55] (0.05429*\pw,-0.065)--(0.05429*\pw,0.065);
  \draw[black!55] (0.31429*\pw,-0.065)--(0.31429*\pw,0.065);
  \fill (0.06286*\pw,0.000) circle (0.05);
  \fill[p2blue] (0.02714*\pw,0.000) circle (0.05);
  \fill[p2purple] (0.03571*\pw,0.000) circle (0.05);
  \node[font=\tiny,anchor=east] at (-0.12,-0.400) {Cups: Large blue cups};
  \draw[capred,line width=0.9pt] (0.31429*\pw,-0.515)--(0.31429*\pw,-0.285);
  \draw[black!55] (0.04857*\pw,-0.400)--(0.09571*\pw,-0.400);
  \draw[black!55] (0.04857*\pw,-0.465)--(0.04857*\pw,-0.335);
  \draw[black!55] (0.09571*\pw,-0.465)--(0.09571*\pw,-0.335);
  \fill (0.08714*\pw,-0.400) circle (0.05);
  \fill[p2blue] (0.02571*\pw,-0.400) circle (0.05);
  \fill[p2purple] (0.04143*\pw,-0.400) circle (0.05);
  \node[font=\tiny,anchor=east] at (-0.12,-0.800) {Cups: Yellow and Green Cups};
  \draw[capred,line width=0.9pt] (0.31429*\pw,-0.915)--(0.31429*\pw,-0.685);
  \draw[black!55] (0.05429*\pw,-0.800)--(0.14143*\pw,-0.800);
  \draw[black!55] (0.05429*\pw,-0.865)--(0.05429*\pw,-0.735);
  \draw[black!55] (0.14143*\pw,-0.865)--(0.14143*\pw,-0.735);
  \fill (0.10286*\pw,-0.800) circle (0.05);
  \fill[p2blue] (0.02571*\pw,-0.800) circle (0.05);
  \fill[p2purple] (0.04286*\pw,-0.800) circle (0.05);
  \node[font=\tiny,anchor=east] at (-0.12,-1.200) {Cups: Three Cups};
  \draw[capred,line width=0.9pt] (0.94286*\pw,-1.315)--(0.94286*\pw,-1.085);
  \draw[black!55] (0.11429*\pw,-1.200)--(0.20857*\pw,-1.200);
  \draw[black!55] (0.11429*\pw,-1.265)--(0.11429*\pw,-1.135);
  \draw[black!55] (0.20857*\pw,-1.265)--(0.20857*\pw,-1.135);
  \fill (0.18143*\pw,-1.200) circle (0.05);
  \fill[p2blue] (0.06714*\pw,-1.200) circle (0.05);
  \fill[p2purple] (0.07571*\pw,-1.200) circle (0.05);
  \node[font=\tiny,anchor=east] at (-0.12,-1.600) {Cups: Sort Cups};
  \draw[black!55] (0.17143*\pw,-1.600)--(0.22571*\pw,-1.600);
  \draw[black!55] (0.17143*\pw,-1.665)--(0.17143*\pw,-1.535);
  \draw[black!55] (0.22571*\pw,-1.665)--(0.22571*\pw,-1.535);
  \fill (0.19643*\pw,-1.600) circle (0.05);
  \fill[p2blue] (0.08571*\pw,-1.600) circle (0.05);
  \fill[p2purple] (0.11429*\pw,-1.600) circle (0.05);
  \node[font=\tiny,anchor=east] at (-0.12,-2.000) {Sorter: Cube};
  \draw[capred,line width=0.9pt] (0.97714*\pw,-2.115)--(0.97714*\pw,-1.885);
  \draw[black!55] (0.18429*\pw,-2.000)--(0.86857*\pw,-2.000);
  \draw[black!55] (0.18429*\pw,-2.065)--(0.18429*\pw,-1.935);
  \draw[black!55] (0.86857*\pw,-2.065)--(0.86857*\pw,-1.935);
  \fill (0.41143*\pw,-2.000) circle (0.05);
  \fill[p2blue] (0.23571*\pw,-2.000) circle (0.05);
  \fill[p2purple] (0.26571*\pw,-2.000) circle (0.05);
  \node[font=\tiny,anchor=east] at (-0.12,-2.400) {Sorter: Prism};
  \draw[capred,line width=0.9pt] (0.97714*\pw,-2.515)--(0.97714*\pw,-2.285);
  \draw[black!55] (0.16857*\pw,-2.400)--(0.96000*\pw,-2.400);
  \draw[black!55] (0.16857*\pw,-2.465)--(0.16857*\pw,-2.335);
  \draw[black!55] (0.96000*\pw,-2.465)--(0.96000*\pw,-2.335);
  \fill (0.46286*\pw,-2.400) circle (0.05);
  \fill[p2blue] (0.13857*\pw,-2.400) circle (0.05);
  \node[font=\tiny,anchor=east] at (-0.12,-2.800) {Sorter: Beam};
  \draw[capred,line width=0.9pt] (0.97714*\pw,-2.915)--(0.97714*\pw,-2.685);
  \draw[black!55] (0.54857*\pw,-2.800)--(0.96429*\pw,-2.800);
  \draw[black!55] (0.54857*\pw,-2.865)--(0.54857*\pw,-2.735);
  \draw[black!55] (0.96429*\pw,-2.865)--(0.96429*\pw,-2.735);
  \fill (0.60929*\pw,-2.800) circle (0.05);
  \fill[p2purple] (0.11429*\pw,-2.800) circle (0.05);
  \fill[p2blue] (0.27429*\pw,-2.800) circle (0.05);
  \node[font=\tiny,anchor=east] at (-0.12,-3.200) {Rings};
  \draw[black!55] (0.32571*\pw,-3.200)--(0.84286*\pw,-3.200);
  \draw[black!55] (0.32571*\pw,-3.265)--(0.32571*\pw,-3.135);
  \draw[black!55] (0.84286*\pw,-3.265)--(0.84286*\pw,-3.135);
  \fill (0.42500*\pw,-3.200) circle (0.05);
  \fill[p2purple] (0.22286*\pw,-3.200) circle (0.05);
  \fill[p2blue] (0.48857*\pw,-3.200) circle (0.05);
  \end{scope}
  \end{tikzpicture}
  \caption{Per-task duration across the $N{=}5$ first-pass trials: the bar spans
  min to max and the black dot marks the median. Blue and purple dots mark the
  second-pass durations, each run by resuming an already-successful trial of that
  task: blue resumes the fastest first-pass trial, purple the slowest (the prism
  has a single successful trial, drawn blue). The red tick marks the task's time
  cap; the caps for Sort Cups (88\,min) and the Rings (316\,min) fall beyond the
  axis.}
  \label{fig:durations}
\end{figure*}

%% file: sections/conclusion.tex
\section{CONCLUSION}
\label{sec:conclusion}
We asked whether a local open-weight vision-language model can control a robot
and generalize, without new human programming or training, to variations of a
task it has only read about. Building on recent work that puts coding agents on
robots, we held the interface, the model and the skill text fixed and varied
the objects and task requirements withing each game, on physical hardware.
Qwen3.8-27B completed 30 of 45 trials without intervention. It generalized
across color and size in 20 of 25 cup trials, inserted every unseen peg at
least once, and rebuilt the ring tower in three of five trials. Where prior
benchmarks vary the interface or the model and lean on API-served frontier
models \cite{capx2026,alrm2026,aspire2026}, we measure generalization across
object and task variations within each game, using a 27B model that runs on one
workstation and can write its own perception code using classical computer
vision.

In our experiments we encountered three failure modes. First, the agent
misreads the scene from its single wrist camera, taking the square hole for the
rectangular one or a standing prism for a lying one. Second, it anchors itself
on a wrong reading and re-measures instead of re-questioning its assumptions.
This can turns a wrong label into an hour-long trial. Third, it lacks awareness
of its own morphology: the platform guards only the tool center point, so a
rotated tool or a wrongly opened gripper can pass every check and end in a
protective stop. Eight of the eleven failures ended in such a stop, one in an
emergency stop. Furthermore, the system is not yet fast enough. A cup task takes four to
fourteen minutes, a sorter task half an hour to an hour, and 80\% of the models
output tokens is reasoning, not the robot moving. The time is bimodal per task:
a trial that gets its first placement right runs in minutes, one that does not
spends its remaining budget in the re-measure loop. A second pass on a solved
task cuts tool calls in half and output tokens to a third, furthermore the fast
trials stay fast on resume. When deploying a local coding agent, it is therefore
beneficial to have a pre-deployment period where the agent learns the task.

In general, a local coding agent shows promise for real-world applications. It
successfully generalized over several object characteristics and task
variations. However, bridging the gap consists of solving the identified
failure modes and making the system faster. We believe there are two
interesting next steps. The first is adding sensory information to remedy both
perception and morphology awareness. The second is further exploring the
presence of a learning and skill consolidation phase.